\documentclass[10pt,twocolumn,letterpaper]{article}

\usepackage{wacv}
\usepackage{times}
\usepackage{epsfig}
\usepackage{graphicx}
\usepackage{amsmath}
\usepackage{amssymb}
\usepackage{multirow}
\usepackage{subfigure}

\usepackage{bbm}

\usepackage{authblk}

\wacvfinalcopy 
\ifwacvfinal
\def\assignedStartPage{9876} 
\fi

\ifwacvfinal
\usepackage[breaklinks=true,bookmarks=false]{hyperref}
\else
\usepackage[pagebackref=true,breaklinks=true,colorlinks,bookmarks=false]{hyperref}
\fi

\ifwacvfinal
\else
\fi

\begin{document}

\title{Self-supervised Visual-LiDAR Odometry with Flip Consistency}


\author[1]{Bin Li}
\author[1]{Mu Hu} 
\author[1]{Shuling Wang}
\author[2]{Lianghao Wang }
\author[1]{Xiaojin Gong\thanks{The corresponding author} \thanks{This work was supported in part by the Natural Science Foundation of Zhejiang Province, China under Grant No. LY17F010007 and LY18F010004.}}

\affil[1]{College of Information Science and Electronic Engineering, Zhejiang University, China\\}

\affil[2]{Zhejiang Lab, Hangzhou, China}

\affil[ ]{\tt\small  { $^1$\{3130102392, muhu, 11831041, gongxj\}@zju.edu.cn, $^2$wang\_sunsky@163.com; }}


\maketitle
\begin{abstract}
Most learning-based methods estimate ego-motion by utilizing visual sensors, which suffer from dramatic lighting variations and textureless scenarios. In this paper, we incorporate sparse but accurate depth measurements obtained from lidars to overcome the limitation of visual methods. To this end, we design a self-supervised visual-lidar odometry (Self-VLO) framework. It takes both monocular images and sparse depth maps projected from 3D lidar points as input, and produces pose and depth estimations in an end-to-end learning manner, without using any ground truth labels. To effectively fuse two modalities, we design a two-pathway encoder to extract features from visual and depth images and fuse the encoded features with those in decoders at multiple scales by our fusion module. We also adopt a siamese architecture and design an adaptively weighted flip consistency loss to facilitate the self-supervised learning of our VLO. Experiments on the KITTI odometry benchmark show that the proposed approach outperforms all self-supervised visual or lidar odometries. It also performs better than fully supervised VOs, demonstrating the power of fusion. 

\end{abstract}

\section{Introduction}
\label{sec:intro}
Ego-motion or pose estimation is a key problem in simultaneous localization and mapping (SLAM), which plays an important role in various applications such as autonomous driving\cite{RGBD-SLAM}, 3D reconstruction~\cite{StereoScan}, and augmented reality~\cite{PTAM}. Most traditional methods~\cite{Nister2004VO,RGBD-SLAM,Zhang2014LOAM,Shan2018LEGOLOAM} address this problem by following a feature extraction and matching pipeline, relying on geometry models for optimization. These methods have shown excellent performance, but they depend on hand-crafted modules and lack the ability of adapting to all challenging scenarios. Recently, learning-based techniques~\cite{SfMLearner,GeoNet,Li2020,L3-Net,Cho2020DeepLO} have attracted more and more research interest. Although these methods perform inferior to traditional counterparts, they demonstrate advantages in dealing with tough cases such as rotation-only motion or tracking lost scenarios.

To date, most learning-based methods use visual sensors. These methods, also referred to as deep visual odometries (VOs)~\cite{SfMLearner,Li2019,Li2020}, take advantage of rich information in monocular or stereo images, but are sensitive to lighting conditions. Several works~\cite{LO-Net,L3-Net,Cho2020DeepLO} adopt ranging sensors such as lidars for ego-motion estimation. These sensors can provide accurate depth measurements and are robust to the change of lighting conditions, but their data are sparse. Therefore, visual and ranging sensors are highly complementary so that it is desirable to integrate them to achieve better performance. 
In addition, considering that most autonomous vehicles are equipped with both visual and ranging sensors nowadays, the study of visual-lidar odometries (VLOs) will be of great value for academic research and industrial applications.


In this work, we propose a learning-based method to implement a visual-lidar odometry. In order to effectively fuse visual and lidar information, we first project 3D lidar points to obtain image-aligned sparse depth maps. Then, we design a network composed of a two-pathway encoder to extract features from a monocular image sequence and a sparse depth map sequence, respectively. The extracted features are fused at multiple scales and employed to jointly predict pose and dense depth values. 
Considering the difficulty of obtaining ground-truth labels, we opt to learn our VLO under self-supervision.
To this end, we adopt a siamese network architecture and design a flip consistency loss to facilitate the self-supervised learning.

The main contributions of our work are summarized as follows:
\begin{itemize}
	\item To the best of our knowledge, this work is the first end-to-end learning-based visual-lidar odometry. It inputs monocular images and sparse depth maps, and outputs both depth and pose estimations.
	\item We adopt a siamese network architecture and design a flip consistency loss to learn our VLO under self-supervision, without using any ground-truth pose and depth labels.
	\item Evaluations on the KITTI odometry benchmark~\cite{KITTI} show that our proposed method outperforms all self-supervised visual or lidar odometries. It also performs better than all fully supervised VOs, demonstrating the power of fusion. 
\end{itemize}

\section{Related Work}
\subsection{Visual \& Lidar Odometry}
\textbf{Visual odometry} aims to estimate ego-motion based on images. Traditional VOs can be classified into feature-based or direct methods. The former~\cite{Nister2004VO,PTAM,ORB-SLAM2} follows a standard pipeline including feature extraction and matching, motion estimation, and local/global optimization. The direct methods~\cite{DTAM,LSD-SLAM} minimize a photometric error that takes all pixels into account, assuming that scenes are stationary and of a Lambertian reflectance. However, these assumptions are hard to be fully satisfied so that most direct methods perform inferior to their feature-based counterparts.

Recently, deep learning techniques have been firstly applied to supervised VO~\cite{DeepVO,ESP-VO,GFS-VO,Xue2019Beyond} that requires ground-truth pose labels. Considering the hardness of obtaining ground-truth, more and more works focus on unsupervised VO. The pioneer work, SfMLearner~\cite{SfMLearner}, takes a monocular video sequence as input to jointly learn depth and pose networks by minimizing a photometric loss. Following it, various end-to-end learning methods have been developed. For instance, UnDeepVO\cite{UnDeepVO}, Zhu et al.\cite{zhu2018robustness}, and Zhan et al.\cite{DepthVOFeat} take stereo sequences as training inputs to address the scale ambiguity in monocular VO. SAVO~\cite{Li2019} and Li et al.~\cite{Li2020} adopt recurrent neural networks (RNNs) to encode temporal information. Additional predictions such as optical flow~\cite{GeoNet}, disparity map~\cite{zhu2018robustness}, and motion mask~\cite{UnDeepVO,Depth-from-wild} are also included to address motion or occlusion problem. In addition, some recent works such as DVSO~\cite{DVSO} and D3VO~\cite{D3VO} integrate the end-to-end pose estimation with traditional global optimization modules to boost VO performance further.

\textbf{Lidar odometry} estimates the relative pose between two 3D point cloud frames. Although a lidar provides accurate depth information, the irregular sparsity of point clouds makes LO a challenging problem. Most previous lidar odometry works, such as LOAM~\cite{Zhang2014LOAM}, LeGO-LOAM~\cite{Shan2018LEGOLOAM}, and LOL~\cite{Rozenberszki2020LOL}, are based on traditional pipelines and achieve state-of-the-art performance. Recently, deep learning techniques have also been applied to this task. For instance, L$^3$-Net~\cite{L3-Net} replaces each module in the traditional pipeline with a deep neural network in order to learn from data. DeepVCP~\cite{DeepVCP}, DeepLO~\cite{Cho2020DeepLO} and LO-Net~\cite{LO-Net} design various end-to-end learning frameworks. The former~\cite{DeepVCP} adopts a 3D convolutional neural network (CNN) for feature extraction and matching, while the latter two~\cite{Cho2020DeepLO,LO-Net} project point clouds into 2D maps via a cylindrical or spherical projection and use 2D CNNs for processing. Most learning-based LOs are performed under full supervision, except DeepLO~\cite{Cho2020DeepLO}.

\textbf{Visual-lidar odometry} takes advantage of visual and lidar information for ego-motion estimation. 
Up to now, most of the existing VLOs are based on traditional methods. For instance, V-LOAM~\cite{Zhang2015V-LOAM} integrates the results of VO and LO, both of which adopt the feature-based pipelines, to boost the performance. LIMO~\cite{Graeter2018LIMO} provides a feature-based monocular VO with the scale estimated from lidar data to achieve ego-motion estimation. 
Recently, there are several visual and lidar based methods making use of learning techniques. Yue et al.~\cite{richLiDAR} exploit a deep network to enrich 3D point clouds with the guidance of high-resolution images. But they employ a traditional point cloud registration method NDT~\cite{NDT} to estimate pose. Vitor et al.~\cite{reprojDis} incorporates depth supervision into a self-supervised VO to produce scale-aware pose estimation. But the input of their network only contains visual images.   
In this work, we propose a self-supervised visual-lidar odometry. It takes a sequence of monocular images and sparse depth maps projected from corresponding 3D lidar points as inputs to learn a depth and pose estimation network in an end-to-end manner, without using any ground-truth pose and depth labels.
%

\subsection{Depth Estimation}
Depth estimation works can be classified into depth prediction and depth completion categories, according to the KITTI benchmark~\cite{KITTI}. The former~\cite{Eigen2014} estimates depth from a single color image, while the latter takes a color image and a sparse depth map as the inputs to produce a dense depth map. In contrast to learning-based VOs~\cite{DeepVO,SfMLearner,DepthVOFeat} that input an image sequence, most depth estimation works take a single frame as input. They focus more on the design of network architectures to encode multi-scale contexts~\cite{Li2020msg}, semantics~\cite{Eigen2015} or surface normals~\cite{DeepLidar} for performance improvement. Moreover, most depth estimation methods are based on supervised learning, requiring ground-truth depth maps that are extremely hard to obtain. Recently, unsupervised methods have also been developed. For instance, Ma et al.~\cite{Ma2019StD} implements self-supervised depth completion via also taking a sequence of images and sparse depth maps as inputs. Their work shares the same inputs with ours, but it uses a traditional method to estimate ego-motion between two frames. 

\begin{figure*}[h]
	\centering
	\includegraphics[width=\linewidth]{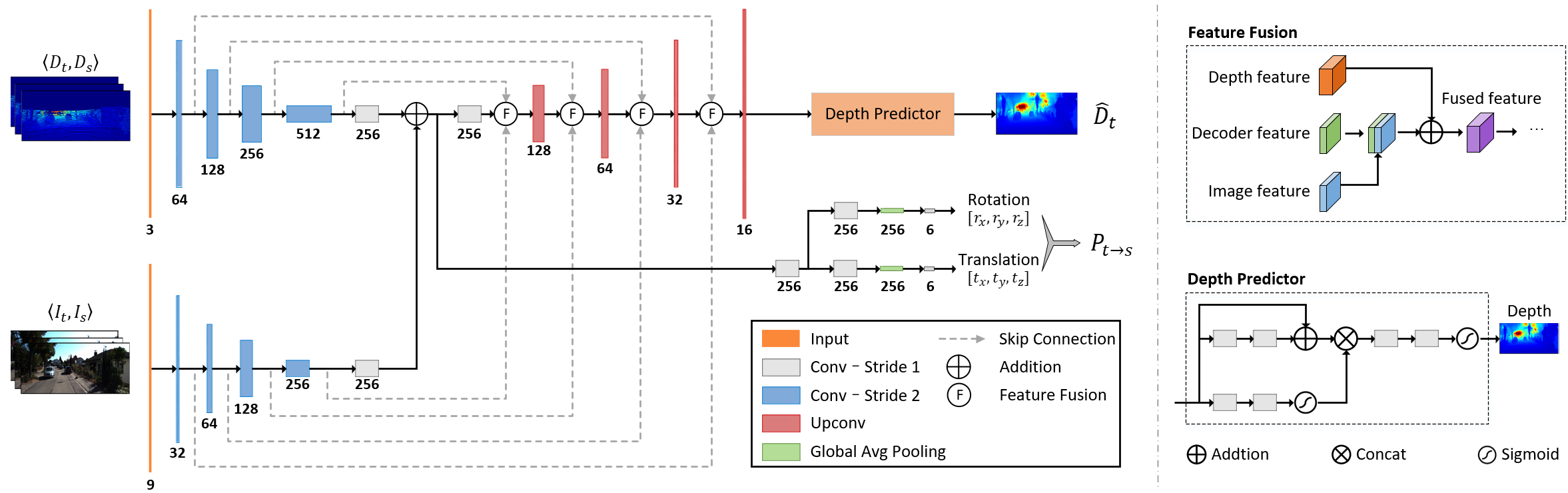}
	\caption{The network architecture of our visual-lidar odometry, referred to as VLONet. It follows an encoder-decoder structure and consists of three major components: a two-pathway encoder, a pose estimator and a depth prediction decoder.}
	\label{fig:sevlo_net}
\end{figure*}

\subsection{Consistency Regularization}
Consistency regularization aims to improve a model's performance via enforcing the consistency between two output predictions. For instance, some stereo matching~\cite{Zbontar2016stereo} and depth estimation~\cite{Chen2019,LRCons,Goldman2019} works take a stereo pair as inputs and place a left-right consistency on their corresponding outputs. Optical flow~\cite{Zou2018DFnet} takes consecutive images as inputs and impose a forward-backward consistency. Recently, the consistency between the outputs of an original image and its transformed (e.g. rescaled, rotated, or flipped) version has been exploited to construct supervision for various self-supervised learning tasks in image classification~\cite{simpleCstL,Guo2019VAC}, object detection~\cite{Jeong2019consistency}, semantic segmentation~\cite{Wang2020SEAM}, etc. In this work, we adopt the flip consistency and design an adaptively weighted consistency loss specifically for self-supervised visual-lidar odometry.   


\section{The Proposed Method}
In this section, we first present the network architecture designed for our visual-lidar odometry. Then we introduce the proposed losses to achieve the self-supervised learning. 

\subsection{Network Architecture}
\textbf{Visual-lidar fusion:} When designing the network architecture, the first problem we need to address is how to fuse visual and lidar modalities. Inspired by recent depth completion works~\cite{DeepLidar,GuideNet}, we adopt an encoder-decoder fusion scheme and design the network as shown in Figure~\ref{fig:sevlo_net}. First, in contrast to deep LOs~\cite{Cho2020DeepLO,LO-Net} that take a cylindrical or spherical projection, we project each 3D point cloud into its aligned image frame and obtain a 2D sparse depth map. Although this projection manner leaves out the 3D points that are not in the camera's field of view, it facilitates the fusion of color and depth information. 

We refer to our visual-lidar odometry network as VLONet. It has a two-pathway encoder. Each takes either an image sequence or a depth map sequence as input. The features extracted in each pathway are fused by a point-wise addition for the prediction of relative pose between two consecutive frames. In addition, the depth and visual features are also fused together with decoder features via a feature fusion module at each scale, as shown in Figure~\ref{fig:sevlo_net}, to predict a dense depth map. In contrast to most deep VOs~\cite{SfMLearner,UnDeepVO,DepthVOFeat} that construct two separate networks for depth and pose prediction, our depth and pose estimation share the same encoder, which can not only reduce the computational cost but also extract more powerful features.

\begin{figure*}[h]
	\centering
	\includegraphics[width=.7\linewidth]{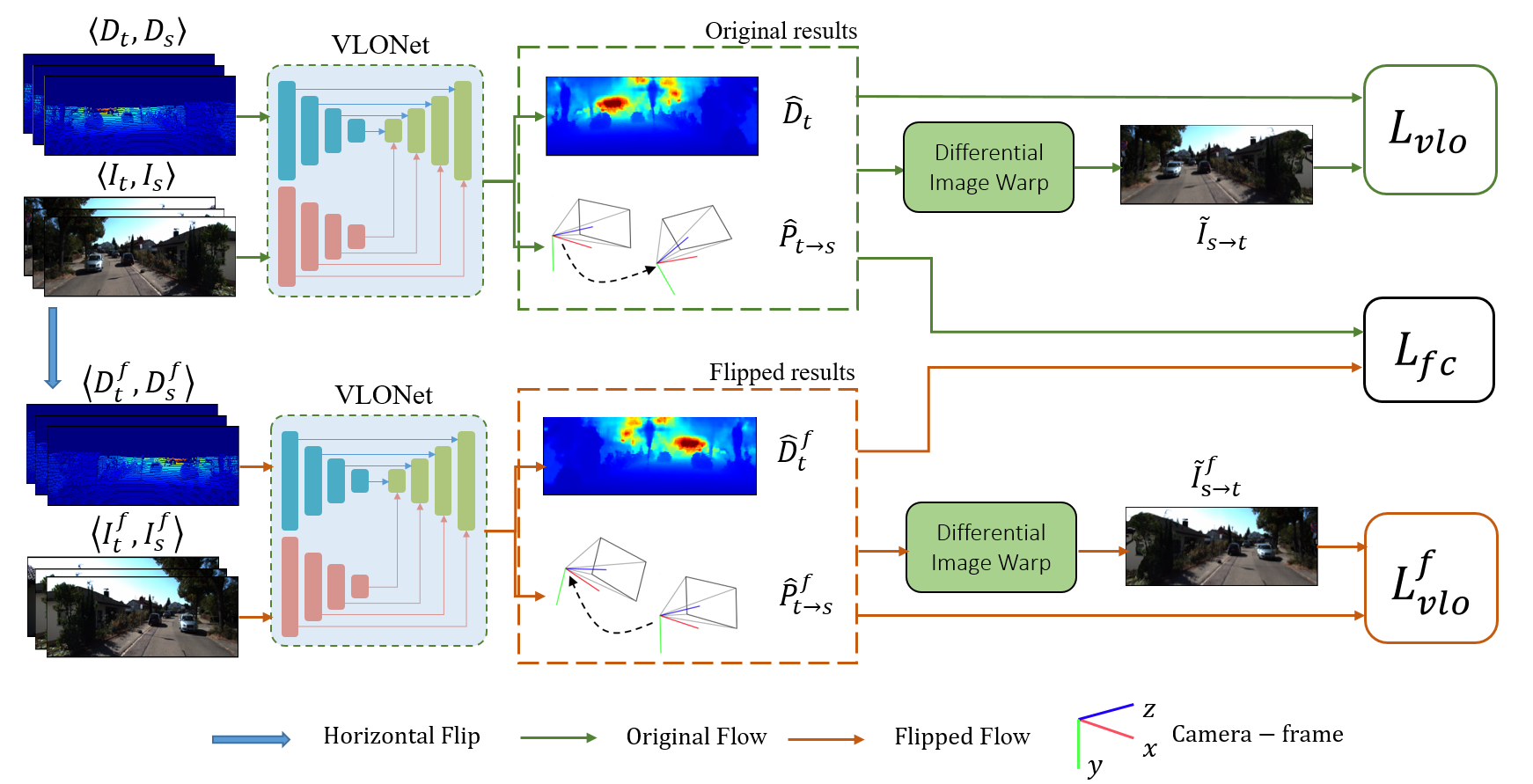}
	\caption{An overview of our Self-VLO. It consists of two VLONets that share the same structures and parameters with each other. One VLONet takes original sequences as input and the other inputs horizontally flipped sequences. Each network is supervised with corresponding depth and pose prediction losses. There is also a flip consistency loss imposed to constrain the outputs of two VLONets.}
	\label{fig:sevlo_overview}
\end{figure*}

\textbf{A siamese architecture for the self-supervised learning:} Inspired by recent unsupervised learning works~\cite{simpleCstL,Guo2019VAC,Jeong2019consistency,Wang2020SEAM}, we adopt a siamese architecture for our self-supervised VLO. More specifically, the architecture consists of two siamese VLONets, as shown in Figure~\ref{fig:sevlo_overview}. One network takes original image and depth sequences as input, and the other takes the horizontally flipped images and depth maps as input. These two networks share the same structure and parameters. Based on this architecture, we can impose a consistency regularization to facilitate our self-supervised learning. This siamese architecture is used for training, while only one VLONet is needed for test.

\subsection{Depth and Pose Prediction Losses}
We first introduce the losses for the prediction of depth and pose in a single VLONet. As no ground-truth labels are available, we adopt a view synthesis loss $\mathcal{L}_{vs}$, together with a depth fidelity loss $\mathcal{L}_{df}$ and a depth smoothness loss $\mathcal{L}_{ds}$ for supervision. Therefore, the entire loss of a single VLONet is defined by
\begin{equation}
\label{eqn:vlo}
\mathcal{L}_{vlo} = \lambda_{vs}\mathcal{L}_{vs} + \lambda_{df}\mathcal{L}_{df} + \lambda_{ds}\mathcal{L}_{ds},
\end{equation}
in which $\lambda_{vs}$, $\lambda_{df}$, and $\lambda_{ds}$ are scaling factors to balance the terms. For the purpose of self-containedness, we briefly introduce each term as follows.

\textbf{View synthesis loss}, which is extensively used in unsupervised VOs~\cite{SfMLearner,UnDeepVO}, measures the difference between a target image and an image synthesized from a source image. Assuming that each sequence contains $2N+1$ frames, we denote the input image sequence by $\{I_{t-N}, \cdots, I_{t-1}, I_t, I_{t+1}, \cdots, I_{t+N}\}$ and the depth map sequence by $\{D_{t-N}, \cdots, D_{t-1}, D_t, D_{t+1}, \cdots, D_{t+N}\}$. The frame $I_t$ and $D_t$ are the target views and the rest are the source views denoted by $I_s$ and $D_s$, $s\in \{t-N, \cdots, t+N\}$, $s\neq t$. The VLONet predicts a dense depth map $\widehat{D}_t$ and a relative pose $\widehat{P}_{t\rightarrow s} = [\hat{t}_x, \hat{t}_y, \hat{t}_z, \hat{r}_x, \hat{r}_y, \hat{r}_z]$ for each target and source pair. Here, $[\hat{t}_x, \hat{t}_y, \hat{t}_z]$ is a 3D translational vector and $[\hat{r}_x, \hat{r}_y, \hat{r}_z]$ is a 3D Euler angle, from which we can construct a $4\times 4$ transformation matrix $\widehat{T}_{t \rightarrow s}$.  

Given a pixel $p_t$ in the target image $I_t$, we can get its corresponding pixel $p_s$ in a source image $I_s$ by the following transform:
\begin{equation}
\label{eqn:pixVS}
p_s = K\widehat{T}_{t \rightarrow s}\widehat{D}_t (p_t)K^{-1}p_t,
\end{equation}
where 
\begin{equation}
K = \left[ \begin{array}{ccc}
f_x & 0 & c_x \\
0 & f_y & c_y \\
0 & 0 & 1
\end{array} \right] 
\end{equation}
is a camera intrinsic matrix provided in training. Further, by taking the differentiable bilinear sampling mechanism~\cite{SfMLearner}, we warp the source image into the target frame to get the synthesized image $\tilde{I}_{s\rightarrow t}$, in which $\tilde{I}_{s\rightarrow t}(p_t) = I_s(\left\langle p_s \right\rangle)$. 

Then, a view synthesis loss is designed to measure the difference between $I_t$ and $\tilde{I}_{s \rightarrow t}$. To this end, we combine a L1 error with a SSIM~\cite{SSIM} and define the loss as follows:
\begin{equation}
\label{eqn:Lvs}
\mathcal{L}_{vs} = \sum_{s} (1-\alpha_{s}) \Vert I_t - \tilde{I}_{s \rightarrow t} \Vert_{1} + \alpha_{s}(1-SSIM( I_t, \tilde{I}_{s \rightarrow t})),
\end{equation}
where $\alpha_{s}$ is a scalar to balance the two terms. According to Equation (\ref{eqn:pixVS}), we know that this loss provides a supervision for both depth and pose estimations.

\textbf{Depth fidelity loss} is to ensure the predicted dense depth map have the same values with the input sparse depth map at these valid pixels. Therefore, this loss is defined by
\begin{equation}
\label{eqn:pwDSup}
\mathcal{L}_{ds}=\Vert M(D_t) \odot  (D_t - \widehat{D}_t) \Vert_{1},	
\end{equation}
in which $M(D_t) = \mathbbm{1} (D_t > 0)$ indicates the valid pixels in the sparse map. Note that the supervision of depth comes from the inputs and no extra ground-truth depth labels are required. Therefore, our method is self-supervised. 

\textbf{Depth smoothness loss} is imposed to improve the quality of the predicted dense depth map. Considering the correlation between color and depth images, we introduce an image-guided loss to smooth depth while keep sharp on object boundaries. This loss is defined by 
\begin{eqnarray}
\label{eqn:depth_smooth}
\mathcal{L}_{sm} &=\sum_p \exp\left(-|\nabla_x I_t(p)|\right)|\nabla_x \widehat{D}_t(p)| \\
&+ \exp\left(-|\nabla_y I_t(p)|\right)|\nabla_y \widehat{D}_t(p)|,
\end{eqnarray}
where $\nabla_x$, $\nabla_y$ denote the gradients. 

\subsection{The Flip Consistency Loss}
By taking advantage of the siamese architecture, we also impose a consistency regularization to facilitate our self-supervised training. It is based on the following assumption~\cite{Wang2020SEAM}: if there is an affine transformation $A(\cdot)$ on the input, the corresponding output is inclined to be equivariant. That is, $G(A(I, D)) = A(G(I, D))$, where $G(\cdot)$ denotes our VLONet function mapping the input to the output. In this work, we only take the horizontal flip into account for simplicity. Therefore, we first derive the relation between two outputs and then design a flip consistency loss to regularize the two outputs. 

Recall that the original input sequences are $\left\langle I_t, I_s\right\rangle$ and $\left\langle D_t, D_s\right\rangle$, and their outputs are $\widehat{D}_t$ and $\widehat{P}_{t\rightarrow s} = [\hat{t}_x, \hat{t}_y, \hat{t}_z, \hat{r}_x, \hat{r}_y, \hat{r}_z]$. We horizontally flip each image and depth frame while keep their temporal order unchanged. The flipped sequences are denoted by $\left\langle I^f_t, I^f_s\right\rangle$ and $\left\langle D^f_t, D^f_s\right\rangle$, and the corresponding outputs are denoted by $\widehat{D}_t^f$ and $\widehat{P}^f_{t \rightarrow s}=[\hat{t}^f_x,~\hat{t}^f_y,~\hat{t}^f_z,~\hat{r}^f_x,~\hat{r}^f_y,~\hat{r}^f_z]$. These two outputs have the following relations:
\begin{equation}
\label{eqn:fc_tsfm}
\left\{
\begin{array}{lr}
\widehat{D}_t^f=flip(\widehat{D}_t) &\\
&\\
\begin{aligned}
\widehat{P}^f_{t \rightarrow s} &= [\hat{t}^f_x,~\hat{t}^f_y,~\hat{t}^f_z,~\hat{r}^f_x,~\hat{r}^f_y,~\hat{r}^f_z] \\
&= [-\hat{t}_x, \hat{t}_y, \hat{t}_z, \hat{r}_x, -\hat{r}_y, -\hat{r}_z] &
\end{aligned}
\end{array}
\right.
\end{equation}
where $flip(\cdot)$ denotes the horizontal flip operation. Note that, when an image is flipped, the intrinsic matrix of the camera is also changed into the following one 
\begin{equation}
\label{eqn:kf}
K^f = \left[ 
\begin{array}{ccc}
f_x & 0 & W - c_x \\
0 & f_y & c_y \\
0 & 0 & 1
\end{array}
\right] 
\end{equation}
where $W$ is the image width. 

Based on the above-mentioned relations, we place consistency constraints on both depth and pose estimation results. Therefore, the flip consistency loss is defined as follows.
\begin{equation}
\label{eqn:fc_loss}
\mathcal{L}_{fc}= \mathcal{L}_{dfc} + \mathcal{L}_{pfc},
\end{equation}
in which 
\begin{equation}
\label{eqn:dfc_loss}
\mathcal{L}_{dfc}=\Vert flip(\widehat{D}_t) - \widehat{D}_t^f\Vert_{1},
\end{equation}
and 
\begin{equation}
\begin{aligned}
\mathcal{L}_{pfc}&=\Vert \hat{t}_x+ \hat{t}_x^f + \hat{t}_y - \hat{t}_y^f + \hat{t}_z - \hat{t}_z^f\Vert_1 \\
&+\alpha_{r} \Vert \hat{r}_x-\hat{r}^f_x + \hat{r}_y + \hat{r}^f_y + \hat{r}_z + \hat{r}^f_z \Vert_1
\end{aligned}
\label{eqn:pfc_loss}
\end{equation}
are two losses to constrain the consistency of the depth and pose estimations, respectively. $\alpha_{r}$ is a scalar to balance the translational error and the rotational error. 

\subsection{The Entire Loss}
In summary, when training with the siamese network architecture, we have two depth and pose estimation losses $\mathcal{L}_{vlo}$ and $\mathcal{L}^f_{vlo}$ , respectively, for the original and flipped sequences, together with a flip consistency loss $\mathcal{L}_{fc}$ to constrain the two outputs. Therefore, the entire loss is 
\begin{equation}
\mathcal{L} = \mathcal{L}_{vlo} + \mathcal{L}^f_{vlo} + \phi \mathcal{L}_{fc}. 
\end{equation}
Here, 
\begin{equation}
\phi = \lambda_{fc} \cdot \exp\left(-\frac{\mathcal{L}_{vs}}{\sigma}\right)
\end{equation}
is an adaptive weight controlled by the view synthesis loss $\mathcal{L}_{vs}$ to balance the prediction losses and the consistency loss. $\lambda_{fc}$ and $\sigma$ are two scalars to ensure the adaptive weight $\phi$ range from 0 to 1. 

The reason to employ this adaptive weighting scheme~\cite{bicyc} is because the consistency loss $\mathcal{L}_{fc}$ is prone to result in a trivial solution with all zeros in depth and pose values. Involving the consistency loss permaturely may make the model parameters oscillate with time when the depth and pose estimates are incorrect, leading to a slow convergence at the early epochs. Therefore, we use this adaptive weighting scheme to vary the weight during training time by letting $\phi$ be inversely proportional to the $\mathcal{L}_{vs}$. $\phi$ is small when $\mathcal{L}_{vs}$ is large and $\phi$ will be close to 1 when $\mathcal{L}_{vs}$ goes to 0 as the training converges.

\begin{table*}
\begin{center}
	\resizebox{.7\textwidth}{!}{
		\begin{tabular}{c c c c c c c c c c c}
			\hline
			\multirow{2}*{Models} & \multirow{2}*{Inputs} & \multirow{2}*{DA} & \multirow{2}*{SA} & \multirow{2}*{FC} & \multicolumn{2}{c}{Seq.09} & \multicolumn{2}{c}{Seq.10} & \multicolumn{2}{c}{Mean} \\
			\cline{6-11}
			\multicolumn{5}{c}{}                          & $t_{rel}$ & $r_{rel}$ & $t_{rel}$ & $r_{rel}$ & $t_{rel}$ & $r_{rel}$ \\
			\hline	   				   
			VLO1       & M+L &        &         &          &   4.33	& 1.72	& 3.30	& 1.40	& 3.82	& 1.56 \\
			VLO2       & M+L &  $\surd$        &         & 		& 3.42 & 1.59	& 4.04	& 1.40	& 3.73	& 1.50 \\
			VLO3      & M+L &  $\surd$	&   $\surd$      &     & 3.21 & 1.33	& 3.56	& \textbf{1.14}	& 3.39	& 1.24 \\
			
		VLO4                & M+L & $\surd$ & $\surd$ & $\surd$ & \textbf{2.58} & \textbf{1.13}	& \textbf{2.67}  & 1.28 & \textbf{2.62} & \textbf{1.21} \\
			\hline
			VO1  & M   &          &         &       & 11.34	& 3.15	& 16.70	& 5.04	& 14.02	& 4.10 \\
			VO2  & M   & $\surd$  &         &       & 9.83	& 3.53	& 14.92	& 4.43	& 12.38	& 3.98 \\
			VO3  & M   & $\surd$  & $\surd$ &       & 8.30	& 2.56	& 15.90	& 4.92	& 12.10	& 3.74 \\
			VO4 & M   & $\surd$  & $\surd$ & $\surd$ & \textbf{7.00}	& \textbf{2.41}	& \textbf{11.74}	& \textbf{3.33}	& \textbf{9.37}	& \textbf{2.87} \\
			\hline
	\end{tabular}}	
	\caption{Comparison of the proposed method and its variants. Here, M denotes the input taking only monocular image sequences, M+L is the input taking both monocular images and depth maps projected from lidar points. DA denotes the data augmentation strategy, SA is the siamese architecture, and FC is the flip consistency loss. $t_{rel}$ is the translational error (\%) and $r_{rel}$ is the rotational error (deg/100m). The best results of VLOs or VOs are marked in bold.}
	\label{tab:exp_ablation}
	\end{center}
\end{table*}

\section{Experiments}
\subsection{Experiment Setting}
\textbf{Dataset and evaluation metrics.} We evaluate our self-supervised visual-lidar odometry and its variants on the KITTI odometry benchmark~\cite{KITTI}. As the common practice~\cite{SfMLearner,UnDeepVO,zhu2018robustness}, we take sequence 00-08 for training and test on sequence 09 and 10. To evaluate the pose estimation results, we compute the average translational error (\%) and rotational error (deg/100m) on all possible sub-sequences of length $(100, 200, \cdots, 800)$ meters, following the official criteria provided in the KITTI benchmark .

\textbf{Implementation details.} We implement the proposed model based on the PyTorch~\cite{PyTorch} framework. In all experiments, the length of each input sequence is 3. Each image or depth map is resized into $192 \times 624$, which is half of the original resolution, to save computational cost. We train our full model using a single NVIDIA GTX 2080Ti. The full model takes 50 hours or so for training. It infers both depth and pose at a rate of 40Hz during test time.

The details of our network, including the operations and channel size, are marked in Figure~\ref{fig:sevlo_net}. In addition, we use batch normalization~\cite{BatchNorm} and ReLU activation for all convolutional layers except those in the prediction layers. During training, we use the Adam optimizer~\cite{adam} with $\beta_1=0.9$, $\beta_2=0.999$ and mini-batch size of 4 to train the network for 180K iterations. The initial learning rate starts from 0.0002 and decreases by half for every 70K iterations. We empirically set the hyper-parameters as follows: $\lambda_{vs}=2.0$, $\lambda_{df}=0.2$, $\lambda_{ds}=40.0$, $\alpha_{s}=0.85$, and $\alpha_{r}=2.0$. In the adaptive weight $\phi$, we set the scaling factors $\lambda_{fc}=2.5$ and $\sigma = 0.2$.

\begin{figure}[h]
	\centering
	\subfigure[Without DA]{
		\includegraphics[width=.2\textwidth]{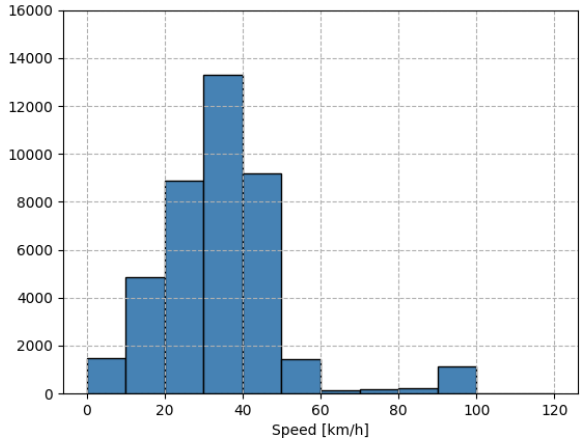}
	}
	\subfigure[With DA]{
		\includegraphics[width=.2\textwidth]{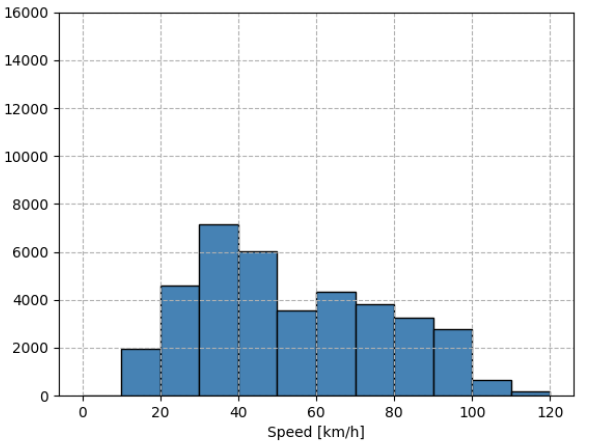}
	}
	\caption{The distributions of the speed between two consecutive frames in 3-frame snippets sampled from the KITTI odometry benchmark (Seq. 00-08). DA refers to the data augmentation strategy designed in our work.}
	\label{fig:sevlo_speed_dist}
\end{figure}

\textbf{Data augmentation.} In our network, each input sequence is a 3-frame snippet. If we only select consecutive frames as inputs, then the low-speed snippets may take the majority while the high-speed snippets take a very small portion, as shown in Figure~\ref{fig:sevlo_speed_dist}(a). This imbalanced distribution makes a model perform poorly at high-speed scenarios. To address this issue, we opt to enlarge the sampling interval by sampling 3 frames, respectively, at time $t-2$, $t$, and $t+2$ with a probability $p = 0.6$. By this means, the speed distribution of two consecutive frames in 3-frame snippets is more balanced, as shown in Figure~\ref{fig:sevlo_speed_dist}(b). In addition, we also observe that extremely low-speed snippets (e.g. the speed between two frames is slower than 10km/h) may hurt the performance. The reason is that the geometry model of pose estimation degenerates when two frames are too close to each other. Therefore, we leave out these snippets during training. Except these, no other data augmentation techniques such as random scaling, cropping or flipping are adopted. 

\subsection{Ablation Studies}
We first conduct a series of experiments to investigate the effectiveness of each component proposed in our model. To this end, we test the following four model variants: 1) VLO1: the model using only a single VLONet and without our data augmentation (DA) strategy; 2) VLO2: the model using one VLONet; 3) VLO3: the model using the siamese architecture composed of two VLONets but without the flip consistency loss; 4) VLO4: the full model. All of the latter three models are trained with DA. In addition, to validate the effectiveness of visual-lidar fusion, we also investigate four visual models, denoted by VO1-4, which are respectively corresponding to VLO1-4 but take only monocular images as input.

\begin{figure}[h]
	\centering
	\subfigure[Seq. 09]{
		\includegraphics[width=.2\textwidth]{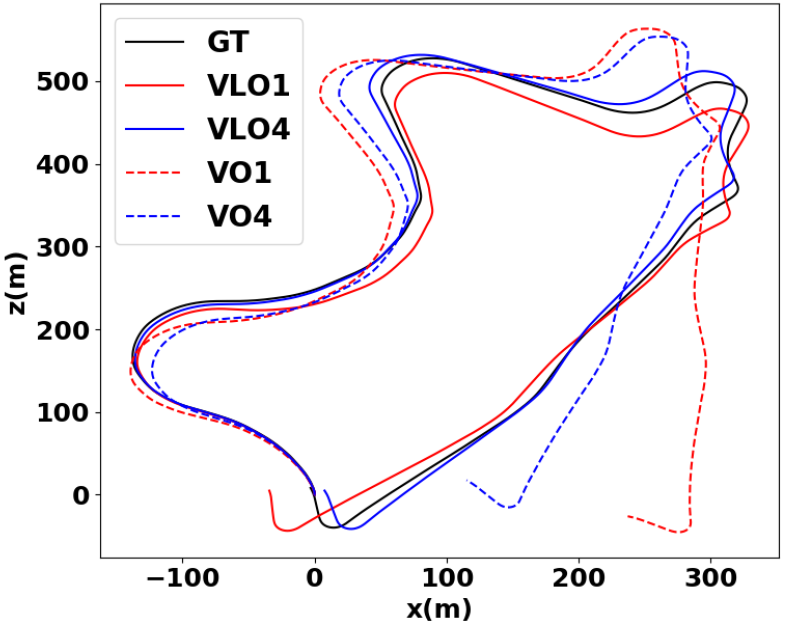}
	}
	\subfigure[Seq. 10]{
		\includegraphics[width=.2\textwidth]{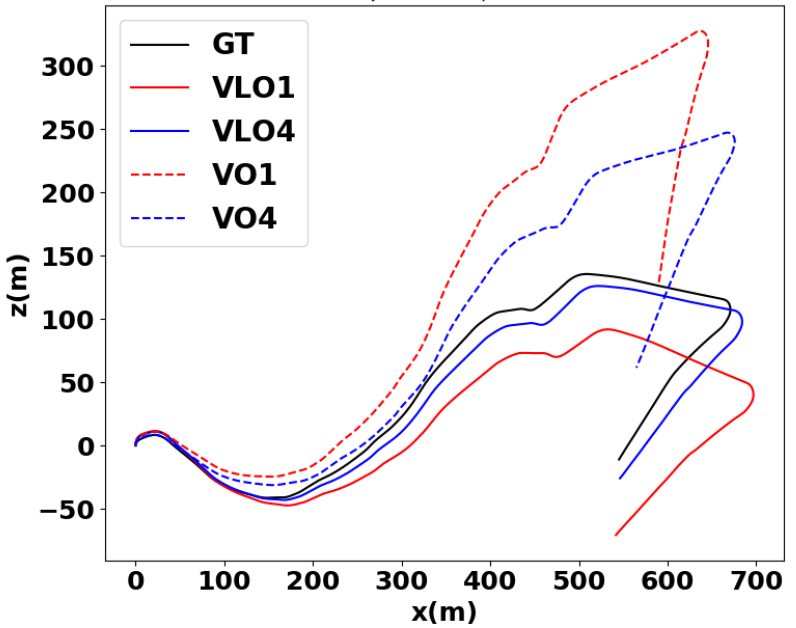}
	}
	\caption{Trajectories of sequence 09 and 10.}
	\label{fig:sevlo_traj_comp}
\end{figure}

Table~\ref{tab:exp_ablation} presents the pose estimation results of all variants. Note that the VO models take only monocular images as input that lead to a scale ambiguity problem, we recover the scales by a post-processing step as done in~\cite{SfMLearner,UnDeepVO}. When comparing each VLO model with the corresponding VO counterpart, we observe that the fusion of visual and lidar information improves the pose estimation performance dramatically. In addition, all proposed components including the data augmentation strategy, the siamese architecture, and the flip consistency loss, gradually boost the performance in both VLOs and VOs, demonstrating their effectiveness. Figure~\ref{fig:sevlo_traj_comp} presents the trajectories of sequence 09 and 10, produced by four model variants. As shown in the figure, both VLO1 and VLO4 generate the trajectories much closer to the ground truth than the VO models, and the full model VLO4 is the closest.

\begin{figure*}
	\centering
	\includegraphics[width=.7\linewidth]{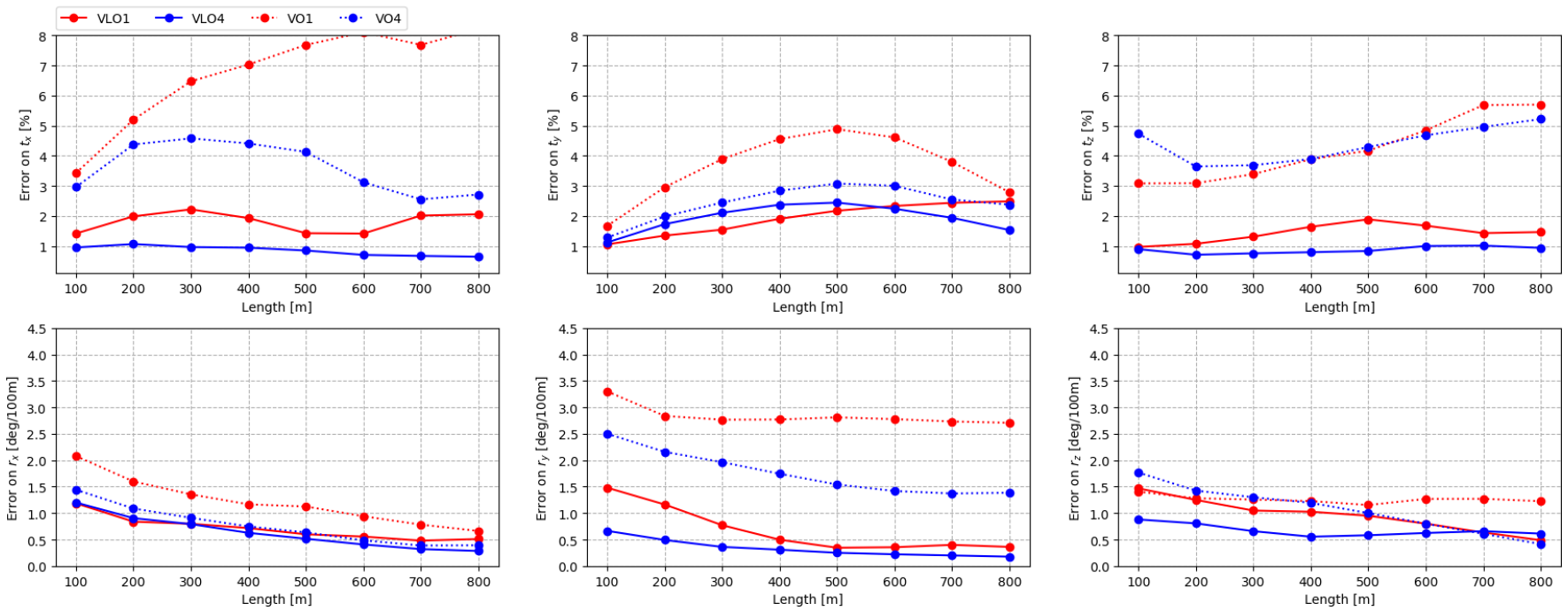}
	\caption{Translational and rotational errors along each axis, averaged on sub-sequences of sequence 09 with different length. Top row: the averaged translational errors along $X$, $Y$, $Z$ axis, respectively. Bottom row: the averaged rotational errors along pitch, yaw, and roll.
}
	\label{fig:sevlo_sp_err}
\end{figure*}

\begin{table*}
	\begin{center}
		\resizebox{.8\textwidth}{!}{
			\begin{tabular}{c l c c  c c c c c c}
				\hline
				\multirow{2}*{ } & \multirow{2}*{Method}  & \multirow{2}*{Inputs} & \multirow{2}*{Supervision} &  \multicolumn{2}{c}{Seq.09} & \multicolumn{2}{c}{Seq.10} & \multicolumn{2}{c}{Mean(09-10)}\\ 
				\cline{5-10}
				\multicolumn{4}{c}{ }   & $t_{rel}$ & $r_{rel}$ & $t_{rel}$ & $r_{rel}$  & $t_{rel}$ & $r_{rel}$\\
				\hline
				\multirow{5}*{\rotatebox{90}{\underline{~Supervised~}}} 
				& DeepVO\cite{DeepVO}     & Mono       & Pose      & -         & -         & 8.11      & 8.83       & 8.11      & 8.83    \\
				& ESP-VO\cite{ESP-VO}     & Mono       & Pose      & -         & -         & 9.77      & 10.2       & 9.77      & 10.2    \\
				& GFS-VO\cite{GFS-VO}     & Mono       & Pose      & -         & -         & 6.32      & 2.33       & 6.32      & 2.33    \\
				& Xue et al.\cite{Xue2019Beyond}  & Mono & Pose    & -         & -         & 3.94      & 1.72       & 3.94     &1.72    \\
				& LO-Net\cite{LO-Net}  & LiDAR          & Pose     & \textbf{1.37}         & \textbf{0.58}         & \textbf{1.80}      & \textbf{0.93}       & \textbf{1.59}      & \textbf{0.76}    \\
				\hline
				\multirow{9}*{\rotatebox{90}{\underline{Self-supervised}}} 
				& SfMLearner\cite{SfMLearner}    & Mono    & -        & 18.77      & 3.21      & 14.33      & 3.30       & 16.55      & 3.26    \\
				& UnDeepVO\cite{UnDeepVO}        & Stereo  & -      & 7.01	  & 3.61      & 10.63      & 4.65       & 8.82	   & 4.13    \\
				& Zhu et al.\cite{zhu2018robustness}  & Stereo & -     & 4.66      & 1.69      & 6.30      & 1.59       & 5.48      & 1.64    \\
				& Zhan et al.\cite{DepthVOFeat}  & Stereo & -      & 11.92      & 3.60      & 12.62      & 3.43       & 12.27      & 3.52    \\
				& Gordon et al.\cite{Depth-from-wild}   & Mono  & - & 3.10      & -         & 5.40      & -          & 4.25      & -       \\
				& SAVO~\cite{Li2019}  & Mono & - &9.52 & 3.64 & 6.45 & 2.41 & 7.99 & 3.03 \\
				& Li et al.~\cite{Li2020}  & Mono & - & 5.89 & 3.34 & 4.79 & \textbf{0.83} & 5.34 & 2.09 \\
				& DeepLO\cite{Cho2020DeepLO}            & LiDAR   & -        & 4.87      & 1.95      & 5.02      & 1.83       & 4.95      & 1.89    \\
				& Self-VLO         & Mono+LiDAR & -      & \textbf{2.58}	& \textbf{1.13}	& \textbf{2.67}	& 1.28	& \textbf{2.62}	& \textbf{1.21}\\
				
				\hline
		\end{tabular}}
		
	\end{center}
	\caption{Comparison of the proposed method with state-of-the-art learning based methods. Note that the self-supervised methods are trained on sequence 00-08. For the supervised methods, \cite{DeepVO,ESP-VO,GFS-VO,Xue2019Beyond} are trained on sequence 00, 02, 08 and 09, and \cite{LO-Net} is trained on sequence 00-06.}
	\label{tab:exp_pose_global}
\end{table*}

In order to take a close look at how the proposed models make improvements, Figure~\ref{fig:sevlo_sp_err} also plots the translational and rotational errors along each axis, averaged on sub-sequences of sequence 09 with a length of $(100, 200, \cdots, 800)$ meters. We see that the fusion of visual and lidar information, as shown by VLO1 and VLO4 in Figure~\ref{fig:sevlo_sp_err}, dramatically improves the performance on $t_x$ (along the lateral direction), $t_z$ (along the driving direction), and $r_y$ (yaw), which are the directions dominate a vehicle's motion. Besides, the errors of VLO4 are more stable than the other variants, indicating that the full model is more robust to the change of driving speed and road scenarios.

\subsection{Comparison with State-of-the-arts}
In this section, we compare our proposed full model (named as Self-VLO) with both fully supervised and unsupervised/self-supervised methods. All previous learning-based methods either take visual images or lidar data as input. Therefore, the state-of-the-art fully supervised methods include four visual odometries~\cite{DeepVO,ESP-VO,GFS-VO,Xue2019Beyond} and one lidar odometry~\cite{LO-Net}. The self-supervised methods include seven VOs~\cite{SfMLearner,UnDeepVO,zhu2018robustness,DepthVOFeat,Depth-from-wild,Li2019,Li2020} and one LO~\cite{Cho2020DeepLO}. These visual methods may take monocular or stereo image sequences as input for training, but all of them infer pose from monocular images during test. 

Table~\ref{tab:exp_pose_global} summarizes the comparison results. From it we see that our Self-VLO outperforms all visual based methods, including both fully supervised and self-supervised ones, on test sequences 09 and 10. Moreover, Self-VLO also performs better than the unsupervised LO~\cite{Cho2020DeepLO} although it takes $360^o$ field of view into account. These comparisons demonstrate the effectiveness of our method. 

\section{Conclusions}
In this paper, we have presented a self-supervised visual-lidar odometry. It takes advantage of rich visual information and accurate depth information to improve the performance of ego-motion estimation. We adopt a siamese architecture and design an adaptively weighted flip consistency loss to facilitate the learning of our VLO under self-supervision, requiring no manual annotations. Experiments on KITTI validate the effectiveness of our proposed method.  


%



%

{\small
\bibliographystyle{ieee_fullname}
\bibliography{sevlo_base_v2}
}

\end{document}